\pdfoutput=1

\documentclass[11pt]{article}

\usepackage[]{ACL2023}

\usepackage{times}
\usepackage{latexsym}

\usepackage[T1]{fontenc}

\usepackage[utf8]{inputenc}

\usepackage{microtype}

\usepackage{inconsolata}
\usepackage{booktabs}

\usepackage{xcolor,pifont}
\newcommand*\colourcheck[1]{%
  \expandafter\newcommand\csname #1check\endcsname{\textcolor{#1}{\ding{52}}}%
}
\colourcheck{blue}
\colourcheck{green}
\colourcheck{red}

\newcommand*\colourx[1]{%
  \expandafter\newcommand\csname #1x\endcsname{\textcolor{#1}{\ding{56}}}%
}
\colourx{blue}
\colourx{green}
\colourx{red}

\definecolor{blued}{HTML}{A9C4EB}
\definecolor{redd}{HTML}{F19C99}

\usepackage{ulem}

\usepackage{graphicx}
\usepackage{multirow}
\usepackage{xspace}
\usepackage{fixltx2e}
\usepackage{array}
\usepackage{float}
\usepackage{placeins}
\usepackage{amssymb}

\newcommand{\storywars}{\textsc{StoryWars}\xspace}
\newcommand{\instruct}{\textsc{InstructStory}\xspace}

\title{\storywars: A Dataset and Instruction Tuning Baselines for \\
 Collaborative Story Understanding and Generation}

\author{Yulun Du \and Lydia Chilton \\
  Columbia University \\
  New York City, New York, USA \\
  \texttt{\{yulundu, chilton\}@cs.columbia.edu} \\
  }

\begin{document}
\maketitle
\begin{abstract}
Collaborative stories, which are texts created through the collaborative efforts of multiple authors with different writing styles and intentions, pose unique challenges for NLP models. 
Understanding and generating such stories remains an underexplored area due to the lack of open-domain corpora.
To address this, we introduce \storywars, a new dataset of over 40,000 collaborative stories written by 9,400 different authors from an online platform.  
We design 12 task types, comprising 7 understanding and 5 generation task types, on \storywars, deriving 101 diverse story-related tasks in total as a multi-task benchmark covering all fully-supervised, few-shot, and zero-shot scenarios.
Furthermore, we present our instruction-tuned model, \instruct, for the story tasks showing that instruction tuning, in addition to achieving superior results in zero-shot and few-shot scenarios, can also obtain the best performance on the fully-supervised tasks in \storywars, establishing strong multi-task benchmark performances on \storywars.\footnote{We make our data, code, and models publicly available at \href{https://github.com/ylndu/storywars}{https://github.com/ylndu/storywars}}
\end{abstract}

\section{Introduction}

\begin{figure}[t]
    \centering
    \hspace*{-0.25cm}
    \includegraphics[scale=0.755,trim={0 0.5cm 0 0}]{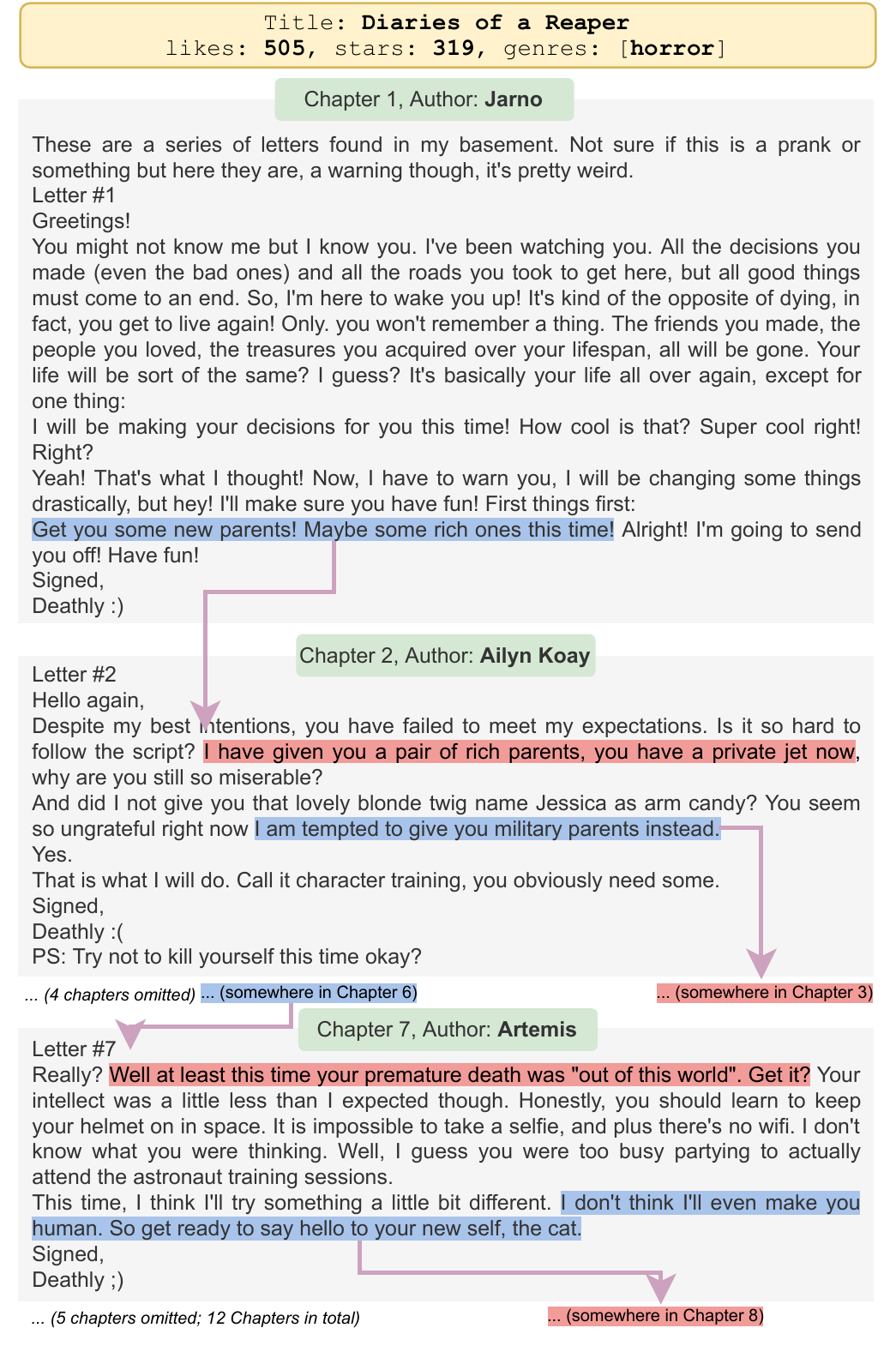}
    \caption{An example story with 12 turns in the \storywars dataset. In each turn, the author leaves a \colorbox{blued}{"floor"} for the next author to \colorbox{redd}{continue collaboratively}.}
    \label{fig:example}
\end{figure}

Storytelling is crucial due to its vital role in human experience, history, and culture dating back to the earliest days of humanity.
Humans possess the unique storytelling ability to structure a sequence of events, whether factual, fictional or a mixture of both, and create a coherent narrative that conveys a big picture while also including intricate details.
Current story generation systems usually mimic this ability by starting with a plot then crafting the story. This can be done by linearly expanding (\citealp{peng2018towards}, \citealp{yao2019plan}, \citealp{MartinAHSHR17}) or hierarchically developing (\citealp{xu-etal-2018-skeleton}, \citealp{fan-etal-2018-hierarchical}, \citealp{fan-etal-2019-strategies}, \citealt{rashkin-etal-2020-plotmachines}, \citealp{goldfarb-tarrant-etal-2020-content}) the story based on the given plot. 
Collaborative storytelling is distinctly challenging because there is no pre-determined plot or story outline of events. Instead, collaborative stories are created through the collective efforts of multiple authors.
Each author contributes a section sequentially, while also attempting to express their own personal intentions within the context of the jointly crafted and jointly owned story.
It is a more challenging problem as it requires not only the ability to generate text, but also the capability to understand the previous context and contributions written by other authors. 

Large Language Models (LLMs) (\citealt{devlin-etal-2019-bert}, \citealp{roberta}, \citealt{xlnet}, \citealt{t5}, \citealt{gpt3}, \citealt{opt}, \citealt{palm}, \citealt{llama}) have demonstrated exceptional performance on various understanding and generation benchmarks, indicating their potential in addressing natural language processing (NLP) challenges related to collaborative storytelling. 
This prompts an intriguing question within the research community: 
\textit{How could LLMs synergize both their understanding and generation capabilities via multitask learning to address the challenges of collaborative storytelling?}

We present \storywars, a dataset of over 40,000 stories gathered from an online collaborative storytelling platform\footnote{\href{www.storywars.net}{www.storywars.net} Unfortunately, the website has closed down by the time of writing this paper. Some stories could be recovered from \href{https://archive.md/sAOOq}{https://archive.md/sAOOq}}.
Figure~\ref{fig:example} shows an example story in the \storywars dataset.
Each story contains rich information including its title, genres given by the initial author, chapters written by different authors, and human ratings including stars and likes.
Each chapter was written by exactly one author and the previous author might leave a \textit{collaborative floor} (\citealp{Coates1997TheCO}) for the next author to continue.
Therefore, for a model to generate a continuing chapter, it needs to understand the preceding context, including the title, genres, and the writing styles and intentions of previous authors conveyed in the collaborative floor.

Due to the multitask nature of collaborative storytelling and the rich information of the \storywars, we design 12 task types, including both understanding and generation task types, as a multitask benchmark for an initial probe of collaborative storytelling. 
We follow the task definition from FLAN \citep{flan}, where each task type contains multiple tasks. 
In the end, our benchmark contains 101 tasks in total, split such that it covers all fully-supervised, few-shot, and zero-shot learning application scenarios. 
It is important to note that prevailing multitask NLP benchmarks are either focusing on understanding (e.g. \citealp{wang-etal-2018-glue}, \citealp{superglue}) or generation (e.g. \citealp{gehrmann-etal-2021-gem}, \citealp{genie}, \citealp{liu-etal-2021-glge}) alone, or only a subset of the learning scenarios.
To our knowledge, we are the first to propose a story benchmark that contains both understanding and generation in all three scenarios.

Large language models have been shown to not only be fully-supervised, few-shot, and zero-shot learners but also multitask ones. 
Instruction Tuning (\citealp{flan}, \citealp{t0}, \citealp{flant5}) has been the state-of-the-art approach for zero-shot and few-shot scenarios. 
However, it has not yet been applied in the fully-supervised setting. 
We evaluated Instruction Tuning on the benchmark and we found that in addition to achieving state-of-the-art results in zero-shot and few-shot scenarios, when combined with single-task fine-tuning, Instruction Tuning can surpass single-task fine-tuning alone, resulting in a consistent performance boost of 1.53 points on average for all tasks.

Our contributions are as follows:
\vspace{-0.3em}
\begin{itemize}
\setlength\itemsep{-0.1em}
    \item We introduce a novel collaborative story dataset \storywars that comprises 40k stories written by 9.4k different authors, with rich information such as genres and human ratings, to promote research in the field of collaborative storytelling.
    \item We propose a new benchmark based on \storywars that consists of 7 understanding and 5 generation task types, totaling in 101 tasks for testing the fundamental abilities of LLMs to model collaborative stories. The benchmark covers the fully-supervised, few-shot, and zero-shot scenarios.
    \item We present \instruct, a instruction-tuned model that demonstrates strong performance on the \storywars benchmark in all three learning scenarios. In addition, we show for the first time that we could extend Instruction Tuning with a single-task finetuning stage to achieve superior performance and obtain robust performance boost.
\end{itemize}
\section{Related Work}

\begin{table*}
  \label{dataset}
  \centering
  \begin{tabular}{cccccccc}
    \toprule
       Dataset & \# Stories &
      \# Words  &
      Genres &
      Human  &
      Open-Domain &
      Multi-Turn  &
      User-Gen  \\
       & & per story & & Ratings & & Collab. &  \\
    \midrule
      ROCStories &  98,156 & 88 & \redx  & \redx  & \greencheck  & \redx & \redx  \\
      WritingPrompts & 303,358 & 735 & \redx & \redx  & \greencheck  & \redx & \greencheck  \\
      roleplayerguild & 1,439 & 3,079 & \redx & \redx  & \redx  & \greencheck & \greencheck  \\
      Storium & 5,743 & 19,278 & \redx  & \redx  & \redx & \greencheck & \greencheck  \\
    \midrule
        \storywars & 40,135 & 367 & \greencheck  & \greencheck & \greencheck & \greencheck  & \greencheck \\
    \bottomrule
    \end{tabular}

    \caption{Comparison of our \storywars dataset with previous story datasets.}
    \label{tab:comp}
\end{table*}

\subsection{Story Datasets}
The most popular story datasets that have been widely used by many story generation systems in the past are ROCStories \citep{mostafazadeh-etal-2016-corpus} and WritingPrompts \citep{fan-etal-2018-hierarchical}. 
ROCStories comprises  five-sentence commonsense short stories, and WritingPrompts includes 300k open-domain prompt-story pairs, neither of which are collaboratively written. 
On the other hand, Storium \citep{akoury-etal-2020-storium} and roleplayerguild \citep{louis-sutton-2018-deep}, are collaborative and written by multiple authors in turns, but in a game setting.
The key distinction of our \storywars dataset is that the stories are both collaborative and open-domain. 
For a comparison of these datasets, refer to Table~\ref{tab:comp}.

\subsection{Multitask NLP Benchmarks}
Existing multitask NLP benchmarks tends to focus on evaluating either understanding (\citealp{wang-etal-2018-glue}, \citealp{superglue}) or generation (\citealp{gehrmann-etal-2021-gem}, \citealp{genie}, \citealp{liu-etal-2021-glge}) capabilities of NLP models. 
There are task-specific benchmarks that address both, such as those for dialog~\citep{MehriDialoGLUE2020} and code~\citep{codexglue}.
For the task of storytelling, the LOT benchmark~\citep{guan-etal-2022-lot} focuses on both aspects but is limited to Chinese and has fewer tasks than our proposed \storywars dataset.
BIG-bench~\citep{Srivastava2022BeyondTI}, which includes 204 tasks for understanding and generation, only tests zero-shot and few-shot abilities without finetuning.
\storywars provides a benchmark for story understanding and generation with 101 tasks spanning all zero-shot, few-shot, and full-supervised scenarios for various applications.

\subsection{Multitask NLP and Instruction Tuning}
Current multitask LLMs mainly follow two approaches. 
The first approach involves finetuning, such as with ExT5~\citep{aribandi2022ext} and Muppet~\citep{aghajanyan-etal-2021-muppet}, where the model is made more generalized through multitask finetuning and then fine-tuned again on downstream tasks.
The second approach focuses solely on zero-shot and few-shot performance, with the goal of bridging the gap between finetuning and these performance levels, as seen in FLAN \citep{flan}, T0\citep{t0}, FLAN-T5 \citep{flant5}, and ZeroPrompt \citep{zeroprompt}. 
These models often utilize Instruction Tuning or similar frameworks. 
In this paper, we extend Instruction Tuning's capabilities to achieve superior performance in the full-supervised scenario as well.

\section{Methodology}
\subsection{The \storywars Dataset}
\label{sec:data}
We obtained the \storywars dataset from \href{www.storywars.net}{storywars.net}, an online collaborative storytelling platform where users can pitch ideas and create stories. 
However, once an initial chapter is published, the story becomes part of the Story Wars community and can be contributed to by other users. 
For a continuing chapter to be officially recognized, it must be voted in by other users, resulting in a high quality of stories on the platform.

We scraped and parsed the stories on Story Wars, ending up in obtaining 76k stories. 
We then used FastText~\citep{bojanowski-etal-2017-enriching} language identification to filter for English stories and further cleaned the dataset by removing noisy stories based on GPT-2 perplexity~\citep{gpt2}. 
We also removed stories that are shorter than 30 words or stories with chapters that are shorter than 10 words. 
To further ensure the quality of the dataset, we also remove stories that have very low human ratings, such as likes and stars.

In consideration of ethical issues, we employed OpenAI Content Moderation APIs\footnote{\href{https://beta.openai.com/docs/api-reference/moderations}{https://beta.openai.com/docs/api-reference/moderations}} and the Detoxify\footnote{\href{https://github.com/unitaryai/detoxify}{https://github.com/unitaryai/detoxify}} toxicity classifier to identify and remove potentially harmful content, such as toxicity, obscenity/sexual content, threats, insults, identity hate, and self-harm posts from the dataset. 
Furthermore, to safeguard user privacy, we replaced all URLs, email addresses, and phone numbers with special tokens <URL>, <EMAIL>, and <PHONE>.

After thorough data cleaning, we obtained a final dataset of 40,135 stories written by 9,494 authors. 
Due to the fact that the long tail of genres is very noisy, we made the simplifying assumption that each story contains a single dominant genre, if any. 
Each story in the dataset was structured with several key elements, including a title, a genre (which could be empty), the numbers of likes and stars received, the authors and the corresponding chapters. 

We denote an arbitrary story in the dataset as $s \in S$, where $S = \{(p, (c_i, a_i)_{i=0}^{t}, g, r_l, r_s)\}$. That is, each story $s_i$ is denoted by a 5-tuple of a title $p$, chapter-author pairs $(c_i, a_i)$ of $t$ turns, a genre $g$, a likes rating $r_l$, and a stars rating $r_s$.

\subsection{The Multitask Benchmark}

\subsubsection{Story Understanding Tasks}

\noindent
\textbf{Genre Classification} 
Understanding the genre of a story is essential for collaborative storytelling models to comprehend the context. 
The genre classification task involves identifying the genre of a story. 
This task can be formulated as a binary text classification problem, where given a story, the task is to predict whether it belongs to a specific genre $g$. 
This can be represented as $g = f(c_1, c_2, ..., c_t)$.

\noindent
\textbf{Authorship Attribution} 
Identifying the author of a text is a crucial step in understanding the writing style of an individual. 
Authorship attribution, traditionally, is the task of determining the author of a given text. 
In this paper, we formulate the task of authorship attribution as identifying the author of a specific chapter, represented as $a = f(c)$.

\noindent
\textbf{Authorship Verification} 
Authorship Verification, in contrast to author attribution, is the task of determining whether two texts have been written by the same author by comparing their writing styles. 
The task is represented as $y = f(c_i, c_j)$, where y is a binary variable.

\noindent
\textbf{Connectivity Inference} 
Understanding the chapter shifts in long-range stories can be a beneficial ability for collaborative storytelling. 
Following \citet{sun-etal-2022-chapterbreak},  we also include the connectivity inference task, where the goal is to determine whether two given chapters are consecutive in a story. The task is represented as $y = f(c_n, c_m)$.

\noindent
\textbf{Temporal Inference} 
Inspired from the Connectivity Inference task, we also aim to evaluate a model's ability to understand the temporal relationships between chapters in a story. 
The Temporal Inference task involves determining whether two chapters in the same story are in the correct chronological order. 
For example, $(c_i, c_{i+1})$ and $(c_i, c_{i+5})$ would be considered positive instances, while $(c_{i+5}, c_i)$ would not. 
The task is represented as $y = f(c_n, c_m)$, where y is a binary variable.

\noindent
\textbf{Story Scoring}
Understanding human ratings of a story is crucial for generating texts that align with human preferences. 
Many dialog-related applications rely on human labelers to rate texts based on different criteria, e.g. LAMDA~\citep{lamda}. 
Since \storywars contains human ratings in the form of likes and stars, we propose to include a regression task for story scoring as a task type. 
We follow \citet{t5} and normalize the story ratings to a range from 0-10, with rounded scores to the nearest increment of 0.1, and convert the float to string.
Given a rating score, such as $r_l$, the task is represented as $r_l = f(c_1, c_2, ..., c_t)$.

\noindent
\textbf{Story Segmentation} 
Although stories are already divided into chapters, it is still possible to evaluate models' ability to identify chapter boundaries within a story, where one chapter concludes and another begins, in order to encourage the model to capture discourse-level information. 
We design the task of story segmentation as $c_1, b_1, c_2, b_2, ..., b_{t-1}, c_t = f(s)$, where $b_i$ is the boundary between two chapters.

\subsubsection{Story Generation Tasks}

\noindent
\textbf{Next Chapter Generation}
The next chapter generation problem is defined as an generation task that takes previous chapters and genre information as input, and then generates the subsequent chapter. 
This is represented as $c_{k+1} = f(c_1, c_2, ..., c_k, g)$.

\noindent
\textbf{Conditional Story Generation}
The conditional story generation problem is defined as an generation task that also takes previous chapters and genre information as input, but then generates the entire continuation of the story until the conclusion instead. 
It further evaluates an NLP model's capability to plan and organize the story.
This is represented as $c_{k+1}, c_{k+2}, ..., c_t = f(c_1, c_2, ..., c_k, g)$.

\noindent
\textbf{Chapter Infilling}
In line with \citet{ippolito-etal-2019-unsupervised}, the chapter infilling task evaluates an NLP model's ability to generate an intermediate chapter given the context of a preceding and subsequent chapter. 
This is represented as $c_k = f(c_{k-1}, c_{k+1})$.

\noindent
\textbf{Global Infilling}
Building on the chapter infilling task, the global infilling problem considers more extensive context information, including both preceding and subsequent chapters. This is represented as $c_k = f(c_1, c_2, ..., c_{k-1}, c_{k+1}, ..., c_t)$.

\noindent
\textbf{Temporal Ordering}
Following \citet{lin-etal-2021-conditional}, we also include a task that unscrambles chapter sequences based on temporal information, except that we simplify the problem by eliminating the requirement for the NLP model to infill masked chapters. 
This is represented as $c_1, c_2, ..., c_{t} = f(permute(c_1, c_2, ..., c_{t}))$.

\begin{table}[t]
\scriptsize
    \centering
    \begin{tabular}{lc|ccc}
    \toprule
\textbf{Task Type}&\textbf{\#Tasks}&\textbf{Train}&\textbf{Dev}&\textbf{Test}\\
\midrule
\midrule
\multicolumn{5}{c}{\textbf{Fully-supervised}}\\
\midrule
Genre Classification & 27 & 2,000 & 250 & 250\\
Author Attribution & 30 & 2,000 & 250 & 250\\
Author Verification  & 1 & 144,000 & 20,925 & 20,925\\
Connectivity Inference  & 1 & 59,402 & 7,521 & 6,963\\
Temporal Inference  & 1 & 84,632 & 9,480 & 8,928\\
Story Scoring  & 2 & 17,046 & 1,485 & 1,484\\
Story Segmentation  & 1 & 17,256 & 1,500 & 1,500\\
\midrule
Next Chapter Generation & 1  & 40,729 & 5,845 & 5,043\\
Conditional Story Generation & 1  & 23,473 & 4,345 & 3,543\\
Chapter Infilling & 1  & 23,473 & 4,345 & 3,543\\
Global Infilling & 1  & 23,473 & 4,345 & 3,543\\
Temporal Ordering & 1  & 78,554 & 8,932 & 8,407\\

\midrule
\midrule
\multicolumn{5}{c}{\textbf{Few-shot}}\\
\midrule
Genre Classification & 10  & 32 & 32 & 200\\
\midrule
\midrule
\multicolumn{5}{c}{\textbf{Zero-shot}}\\
\midrule
Genre Classification & 23  & 0 & 0 & 200\\
\bottomrule
    \end{tabular}
    \caption{Task statistics for the \storywars benchmark.}
    \label{tab:tasks}
\end{table}

\subsubsection{The Benchmark}
\noindent
\textbf{Benchmark task statistics}
The 12 task types translate into 101 tasks based on \storywars, with 96 understanding tasks and 5 generation tasks. 
It is worth noting that the majority of the understanding tasks are genre classification tasks (60) and author attribution tasks (30). 
Out of the 60 genre classification tasks, we split them into 27 fully-supervised, 10 few-shot, and 23 zero-shot datasets, according to the genre frequency so that the split closely aligns with realistic data distribution. 
For the fully-supervised and few-shot tasks, we divided the data into training, dev, and test sets. 
For the zero-shot tasks, we used all the data as a test set by sampling. 
The remaining task types were used for fully-supervised scenarios. 
It is important to mention that all of the data in the fully-supervised, few-shot, and zero-shot scenarios are disjoint to prevent data leakage. 
The overall task data statistics can be found in the Table~\ref{tab:tasks}.

\noindent
\textbf{Evaluation metrics}
For the genre classification, author attribution, author verification, temporal inference, and connectivity inference tasks, we use F-1 score as the evaluation metric, due to the imbalance nature of the task data. 
For the story scoring tasks, in line with \citet{t5} for regression tasks, we use Spearman correlation coefficients as the evaluation metric, because it measures monotonic relationships.
For the story segmentation task, we use Boundary Similarity \citep{fournier-2013-evaluating} as the evaluation metric. 
For the generation tasks, following the suggestions introduced in \citet{chhun-etal-2022-human}, \citet{qin-etal-2019-counterfactual}, and \citet{nareor}, we use BERTScore~\citep{Zhang*2020BERTScore:} as the evaluation metric, as it has been shown by \citet{chhun-etal-2022-human} to have better correlation with human evaluation at both the story-level and system-level for story generation systems than other automatic metrics including frequently-used BLEU~\citep{papineni-etal-2002-bleu} and ROUGE~\citep{lin-2004-rouge}. 
Also, \citet{nareor} points out that in the narrative reordering problem, similar to our temporal ordering task, BERTScore also correlates quite well with human evaluations. 
We recognize that there is currently no widely accepted or reliable automatic evaluation metric in the field of story generation, and the use of automatic evaluation in this field is often criticized. 
However, for the purpose of fast and fair comparison, we chose to follow previous work and use the current best available metric, even though we acknowledge that it may not be perfect. 

For evaluating the model performance, we calculate the macro-average of the performance on all tasks within each task type, this allows us to compare models across different task types. The metrics for understanding, generation, and overall performance are determined by the macro-average of the scores across the corresponding task types.

\subsection{The \instruct Framework}
The main goal of instruction tuning is to evaluate the performance of unseen tasks in zero-shot and few-shot learning scenarios, and to show that it can improve the gap between zero-shot and fully-supervised learning performances.
Additionally, we are interested in how instruction tuning can improve the performance of fully-supervised tasks.

To accomplish our goal, we propose a two-stage training approach called \instruct. In the first stage, we use instruction tuning as a form of \textit{pre-finetuning}~\citet{aghajanyan-etal-2021-muppet}. 
During this stage, we use instructions instead of task prefixes proposed in Muppet~\citet{aghajanyan-etal-2021-muppet} to enhance the model's ability to generalize to new instructions. 
In the second stage, after instruction tuning with the fully-supervised task mix, we use single-task finetuning to continually train the model for each fully-supervised task. 
We use T5-large-lm-adapt (770m) as the base model for instruction tuning \instruct and all of the training tasks are from the \storywars fully-supervised training split. 
Figure~\ref{fig:instuctstory} illustrates the overall \instruct framework. The instructions we used are included in Appendix~\ref{a1}.

\begin{figure*}[t]
    \centering
    \includegraphics[width=1\textwidth]{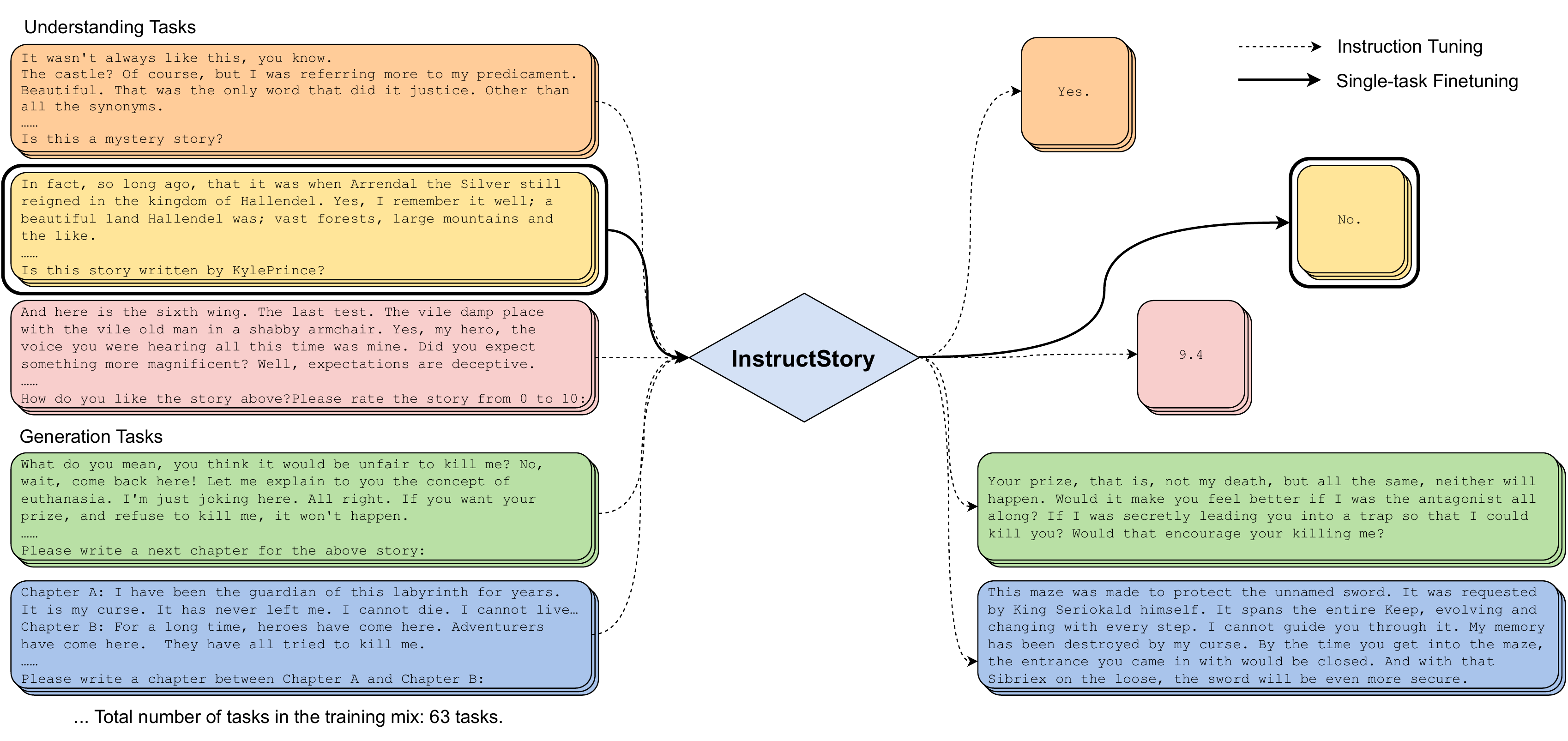}
    \caption{\instruct undergoes a two-stage training process. In stage 1 ($\dashrightarrow$), we instruction tune the model on 63 story tasks to improve generalization to unseen zero-shot and few-shot tasks. In stage 2 ($\rightarrow$), we perform single-task finetuning on each fully-supervised task to optimize performance on specific tasks.}
    \label{fig:instuctstory}
\end{figure*}

\begin{table*}[!ht]
    \small
    \newcolumntype{H}{>{\setbox0=\hbox\bgroup}c<{\egroup}@{}}
     \hspace*{0.1cm}
    \begin{tabular}{ccccccHc} 
    \midrule
         Task Type & Task & \textbf{BERT} & \textbf{RoBERTa} & \textbf{DeBERTa} & \textbf{T5} & \textbf{FLAN-T5} & \textbf{InstructStory}  \\ 
         \midrule
\multirow{10}{*}{\textbf{Genre Classification}\dag}  & \scriptsize animals & 82.69 & 86.02 & 82.24 & 82.88 & 0  & \textbf{86.79} \\ 
                              ~ & \scriptsize fantasy         & 43.70 & 47.37 & 48.75 & 47.95  & 0 & \textbf{50.98} \\
                              ~ & \scriptsize horror          & 45.67 & 55.64 & \textbf{60.15} & 52.05 & 0  & 53.33 \\ 
                              ~ & \scriptsize war         & 59.77 & 68.97 & 76.00 & 70.59  & 0 & \textbf{78.26} \\
                              ~ & \scriptsize poetry           & 78.90 & \textbf{85.71} & 79.65 & 81.97 & 0  & 84.96 \\
                              ~ & \scriptsize drama         & 42.67 & 45.30 & 46.43 & 44.21 & 0  & 47.40 \\
                              ~ & \scriptsize mystery          & 43.58 & 51.47 & 48.53 & 47.48 & 0  & \textbf{51.97} \\
                              ~ & \scriptsize fanfiction         & 55.28 & 62.26 & \textbf{67.27} & 63.41 & 0  & 66.07 \\
                              ~ & \scriptsize dystopia          & 43.48 & 57.14 & 61.16 & 52.23 & 0  & \textbf{63.55} \\
                              ~ & \scriptsize sci-fi         & 65.42 & 61.07 & \textbf{67.24} & 62.69 & 0  & 66.67 \\
                              \cmidrule(r){2-8}
                              ~ & \scriptsize AVG         & 51.86 & 61.15 & \textbf{62.20} & $60.15$ & $61.62$ & $\underline{61.88}$ \\
                              \midrule
\multirow{10}{*}{\textbf{Author Attribution}\dag}   & \scriptsize aspiringwriter              & 66.67 & \textbf{69.57} & 62.02 & 60.40 & 0  & 67.18 \\ 
~ & \scriptsize sagittarius           & 50.94 & 54.74 & 58.02 & 48.52 & 0  & \textbf{64.81} \\ 
                              ~ & \scriptsize Hope!         & 61.82 & \textbf{81.13} & 62.30 & 56.21 & 0 & 68.22 \\
                              ~ & \scriptsize Shasta          & 52.17 & 55.56 & 58.49 & 37.04 & 0  & \textbf{59.38} \\
                              ~ & \scriptsize Scorpio :)        & 61.82 & \textbf{81.13} & 62.30 & 56.21 & 0 & 68.22 \\
                              ~ & \scriptsize Zed           & 67.27 & 72.94 & \textbf{81.82} & 73.27 & 0  & 78.85 \\
                              ~ & \scriptsize Nathan.N         & 82.61 & 84.78 & 86.00 & 86.32 & 0  & \textbf{87.23} \\
                              ~ & \scriptsize Ellipsis          & 78.85 & \textbf{83.67} & 59.38 & 67.89 & 0 & 78.00 \\
                              ~ & \scriptsize Luke V.	         & 72.09 & 69.77 & 69.23 & 63.24 & 0 & \textbf{73.79} \\
                              ~ & \scriptsize Amelia Rose	   & 50.00 & \textbf{70.10} & 68.57 & 53.62 & 0 & 68.97 \\
                              \cmidrule(r){2-8}
                              ~ & \scriptsize AVG         & 64.52 & \textbf{72.31} & 69.08 & $62.03$ & $57.17$ & $\underline{70.79}$ \\
                              \midrule
\multirow{1}{*}{\textbf{Author Verification}} & \scriptsize author\_verification         & 23.19 & 23.41 & 23.17 & 22.94 & 20.23 & \underline{\textbf{23.57}} \\ \midrule
\multirow{1}{*}{\textbf{Temporal Inference}}    & \scriptsize temporal\_inference            & 72.90 & 77.74 & \textbf{80.18}  & 78.51 & \textbf{79.98} & \underline{79.04} \\ \midrule
\multirow{1}{*}{\textbf{Connectivity Inference}}    & \scriptsize connectivity\_inference           & 65.03 & 62.97 & 67.61 & 67.20 & 67.63 & \underline{\textbf{68.72}} \\ \midrule
\multirow{2}{*}{\textbf{Story Scoring}}   & \scriptsize likes\_scoring                   & 53.54 & \textbf{75.74} & 60.81 & 67.35 & 66.05 & \underline{68.82} \\ & \scriptsize stars\_scoring                   & 55.34 & \textbf{66.60} & 56.02 & 63.15 & 60.01 & \underline{63.26} \\ \midrule
\multirow{1}{*}{\textbf{Story Segmentation}}    & \scriptsize story\_segmentation           & 31.38 & 47.28 & 41.09 & 46.87 & 42.56 & \underline{\textbf{47.33}} \\ \midrule
\multicolumn{2}{c}{\textbf{Understanding AVG}}                                     & 51.90 & 59.43 & 57.39 & 57.56 & 57.56 & \underline{\textbf{59.62}} \\ \midrule 
\midrule
 Task Type & Task & \textbf{GPT2-l} & \textbf{GPT2-m} & \textbf{OPT-350m} & \textbf{T5} & \textbf{FLAN-T5} & \textbf{InstructStory}  \\ \midrule

\multirow{1}{*}{\textbf{Next Chapter Generation}}    & \scriptsize next\_chapter         & 81.35 & 80.90 & \textbf{83.25} & 82.17 & 0 & $\underline{82.43}$ \\ \midrule 
\multirow{1}{*}{\textbf{Conditional Story Generation}}    & \scriptsize conditional      & 79.40 & 79.33 & \textbf{82.39} & 81.10 & 0 & $\underline{81.24}$ \\ \midrule
\multirow{1}{*}{\textbf{Chapter Infilling}}    & \scriptsize chapter\_infilling                & 80.93 & 80.67 & \textbf{82.89} & 82.34  & 0 & $\underline{82.51}$ \\ \midrule
\multirow{1}{*}{\textbf{Global Infilling}} & \scriptsize  global\_infilling                   & 81.49 & 81.30 & \textbf{83.70} & 82.22 & 0 & $\underline{82.44}$ \\ \midrule
\multirow{1}{*}{\textbf{Temporal Ordering}}   & \scriptsize temporal\_ordering                 & 76.49 & 76.33 & 92.77 & 90.08 & 83.30 & $\underline{\textbf{93.14}}$ \\ \midrule

\multicolumn{2}{c}{\textbf{Generation AVG}}   & 79.93 & 79.71 & \textbf{85.00} & 83.58 & 0 & $\underline{84.35}$ \\ \midrule \midrule

\multicolumn{2}{c}{\textbf{Understanding and Generation Overall AVG}}  & - & - & - & 68.40 & - & $\underline{\textbf{69.93}}$ \\
\bottomrule
    \end{tabular}
    \caption{Fully-supervised results of \instruct and other baselines. \textbf{Bold} numbers indicate the best score across all models, and \underline{underlined} numbers indicate cases where \instruct outperforms the T5 baseline. Due to space limits, only 10 random tasks from the task type are shown. Full results can be found in the Appendix~\ref{a3}.}
    \label{tab:main_results}
\end{table*}

\section{Experimental Results}

\subsection{Baselines}
We include several strong baseline models with a comparable number of parameters. 
For understanding tasks, we include \textbf{BERT-large} (345m), \textbf{RoBERTa-large} (354m), and \textbf{DeBERTa-v2-xlarge} (900m) as baselines. 
For generation tasks, we include \textbf{GPT2-medium} (345m), \textbf{GPT2-large} (774m), and \textbf{OPT-350m} as baselines. 
These models all have comparable or near-comparable numbers of parameters. 
To demonstrate the effectiveness of our method, we also include \textbf{T5-large-lm-adapt} (770m) as a baseline model in the overall comparison. 
In addition, for the few-shot and zero-shot scenarios, we include the state-of-the-art instruction tuning model \textbf{FLAN-T5-large}~\citep{flant5} as a comparison baseline.

\begin{table*}[]
	\small
    \centering
    \begin{tabular}{ccccccc}
    \midrule
         task & \textbf{BERT} & \textbf{RoBERTa} & \textbf{DeBERTa} & \textbf{T5} & \textbf{FLAN-T5} & \textbf{InstructStory}  \\ 
         \midrule
\scriptsize wordgames & 59.65 & \textbf{80.90} & 77.27 & 62.40 & 71.05 & 73.68 \\ 
                                \scriptsize rebellion         & 38.38 & 45.87 & 33.33 & 43.24 & \textbf{50.00} & \textbf{50.00} \\
                               \scriptsize mythology           & 47.27 & 59.79 & 61.54 & 62.07 & 66.67 & \textbf{67.33} \\ 
                               \scriptsize future         & 30.00 & 40.00 & 50.90 & 36.23 & 44.86 & \textbf{54.70} \\
                               \scriptsize friendship           & 38.82 & 46.96 & 44.62 & 49.23 & 53.33 & \textbf{55.36} \\
                               \scriptsize fairytale         & 45.93 & 60.32 & 65.52 & 74.07 & 72.09 & \textbf{79.59} \\
                               \scriptsize dreams           & 47.48 & 64.15 & 58.62 & \textbf{78.16} & 71.26 & 76.74 \\
                               \scriptsize crime         & 48.54 & \textbf{66.67} & 36.04 & 65.42 & 62.22 & 65.26 \\
                               \scriptsize change           & 44.00 & \textbf{50.36} & 32.91 & 33.90 & 47.89 & 39.19 \\
                               \scriptsize action         & 38.30 & 40.25 & 36.47 & 41.13 & \textbf{55.10} & 52.54 \\
                              \midrule
                               \scriptsize AVG         & 43.84 & 55.53 & 49.72 & 54.59 & 59.45 & \textbf{61.44} \\
                              \midrule
    \end{tabular}
    \caption{Few-shot benchmark results. \instruct outperforms all other baselines.}
    \label{tab:fewshot}
\end{table*}

\subsection{Experimental Setup}
To train \instruct, we use instruction tuning on T5-large-lm-adapt for 5 epochs using the fully-supervised task mix. We use the Adam optimizer with a learning rate of 5e-5 and a batch size of 64. At each gradient step, examples are randomly sampled from all tasks. The maximum input and target sequence lengths are set to 1024, and any longer inputs or targets will be truncated.

For the fully-supervised learning scenario, both \instruct and all the baselines are fine-tuned on a single task for 10 epochs for each task. 
The best performing checkpoint for each task is chosen based on the performance on its dev set. 
Note that BERT-large, RoBERTa-Large, and DeBERTa-v2-xlarge all have a maximum sequence length of 512, while GPT2-medium and GPT2-Large have a maximum sequence length of 1024 and OPT-350m has a maximum sequence length of 2048. 
We truncate the data instances based on the respective max sequence lengths of the models.

For the few-shot learning scenario, we finetune all the models and use early stopping based on the dev set performance. 
Also, we are unable to use in-context learning demonstrations like in \citet{flant5}, as the story lengths are often too long to fit within the max input sequence length.

For the zero-shot scenarios, we only compare \instruct with T5 and FLAN-T5, as the other baseline models have poor zero-shot performance.

More information about training specifics and hyperparamters can be seen in Appendix~\ref{a2}.

\begin{table}[t]
	\small
    \centering
    \begin{tabular}{cccc}
    \midrule
          task\dag & \textbf{T5} & \textbf{FLAN-T5} & \textbf{InstructStory}  \\ 
         \midrule
\scriptsize reality  & 32.56 & 39.56 & 39.47 \\ 
\scriptsize lies   & 30.22 & 46.34 & 70.33  \\
\scriptsize vampire  & 19.12 & 63.33 & 58.82  \\
\scriptsize surreal  & 31.41 & 33.86 & 46.25  \\
\scriptsize suspense  & 31.82  & 42.77 & 43.68  \\
\scriptsize supernatural & 39.34 & 48.28 & 45.33  \\
\scriptsize family   & 14.88 & 51.16 & 60.00  \\
\scriptsize revenge    & 35.00 & 58.06 & 57.14 \\
\scriptsize crazy  & 30.00 & 42.31 & 43.08   \\
\scriptsize world & 30.63 & 34.92 & 50.75   \\
\midrule
\scriptsize AVG     & 32.09 & 47.79 & \textbf{60.00}   \\
\midrule
    \end{tabular}
    \caption{Zero-shot benchmark results. \instruct out performs T5 and even FLAN-T5. \dag: Due to space limits, we only show 10 random tasks. Full results can be found in Appendix~\ref{a3}.}
    \label{tab:zeroshot}
\end{table}

\subsection{Main Results}
\noindent
\textbf{Fully-supervised Results} 
The fully-supervised results are presented in Table~\ref{tab:main_results}. 
We show that \instruct can achieve a 1.53 point increase in the overall average score compared to the single-task finetuned T5 baseline. 
Additionally, for understanding tasks, \instruct outperforms T5 by 2.06 points. 
When compared to other strong understanding baselines including BERT, RoBERTa, and DeBERTa, \instruct also achieves the best results. 
For generation tasks, \instruct outperforms T5 by 0.77 points. 
It also achieves favorable performance when compared to other strong generation baselines such as GPT2-medium and GPT2-large, although performing a little bit worse than OPT-350m. 
We hypothesize that the difference in performance between OPT-350m and \instruct is due to the base model, specifically the size of the pretraining corpus (35B tokens vs 180B tokens).\citep{opt}

\noindent
\textbf{Few-shot Results}
The few-shot results are shown in Table~\ref{tab:fewshot}. For the few-shot scenario, \instruct achieves the highest score of 61.44, followed by FLAN-T5 which achieved the second highest score of 59.45, outperforming all the T5, BERT, RoBERTa, and DeBERTa baselines. 
This demonstrates that even when instruction-tuned on a different dataset distribution, FLAN-T5 can still achieve competitive results when further fine-tuned for few-shot tasks.

\noindent
\textbf{Zero-shot Results} We can see the zero-shot results in Table~\ref{tab:zeroshot}. In the zero-shot scenario, we compare \instruct with T5 and FLAN-T5, and we can see that \instruct has a significant improvement in zero-shot performance, a 28.08 increase from T5 and a 12.21 increase from FLAN-T5. 
This is expected because our instruction tuning training task mix has a similar, though unseen, data distribution to the zero-shot test sets.

\subsection{Discussions}

\noindent
\textbf{\instruct brings a robust improvement in performance.} By comparing T5 and \instruct in Table~\ref{tab:main_results}, we see that \instruct scores higher than T5 in every task type. The performance gain is consistent across all task types. Even on the task level, \instruct achieves better results than T5 in 24 out of 27 genre classification tasks and 23 out of 30 authorship attribution tasks. 
This indicates that in fully-supervised scenario, one can confidently use the power of instruction tuning to improve performance.



\begin{table}[t]
	\small
    \centering
    \begin{tabular}{ccccc}
    \midrule
           &  \textbf{IS} & \textbf{IS\textsubscript{U}} & \textbf{IS\textsubscript{G}} &  \textbf{T5} \\ 
         \midrule
\scriptsize Fully-sup AVG  & 61.88 & 61.27 & 60.45 & 60.15 \\ 

\scriptsize Few-shot AVG  & 61.44 & 59.83 & 54.95 & 54.59 \\ 

\scriptsize Zero-shot AVG  & 60.00 & 58.41 & 32.31 & 32.09 \\ 
\midrule
    \end{tabular}
    \caption{\instruct vs its variants IS\textsubscript{U} and IS\textsubscript{G}.}
    \label{tab:ablation}
\end{table}

\noindent
\textbf{Ablation: Instruction tuning with both understanding and generation tasks is more effective than instruction tuning with only understanding tasks or only generation tasks.} 
Table~\ref{tab:ablation} illustrates this by comparing the fully-supervised, few-shot, and zero-shot genre classification scores of \instruct, its variants IS\textsubscript{U}, and IS\textsubscript{G}, where IS\textsubscript{U} and IS\textsubscript{G} are instruction tuned with understanding tasks mix and generation tasks mix, separately. 
From the table, we can see that IS > IS\textsubscript{U} > IS\textsubscript{G} > T5 across all zero-shot, few-shot, and fully-supervised learning scenarios, which indicates that instruction tuning with a mix of understanding and generation tasks is better than instruction tuning with only one of them.



\section{Conclusion}

We introduced a novel dataset \storywars and a multitask benchmark for collaborative story understanding and generation. 
Our proposed \instruct model, which leverages instruction tuning as multitask pre-finetuning, outperformed both its single-task finetuning baseline and other strong models on the \storywars benchmark and established strong performance in all zero-shot, few-shot, and fully-supervised learning scenarios. We hope that our newly proposed \storywars dataset will serve as a catalyst for research in the field of collaborative storytelling and inspire further advancements in this area.

\section{Limiations}
Our proposed \instruct method utilizes both single-task finetuning and instruction tuning to achieve good results. However, when finetuned on a new task, the model may suffer from the problem of catastrophic forgetting and lose its multitasking generalization abilities. Recent research by \citet{Scialom2022FinetunedLM} has investigated this issue in instruction-tuned models and proposed a technique called Rehearsal to mitigate it. However, this work primarily focuses on zero-shot scenarios and does not address fully-supervised learning. It would be of interest to explore whether it is possible to finetune on a single task while preserving the model's multitasking abilities and generalization capabilities. We leave this question as an area for future research.

 Additionally, it is important to note that our approach of single-task finetuning for each downstream task results in multiple models being required to be served simultaneously, which can lead to increased computational costs. In practice, this is a trade-off that must be carefully considered, as it requires balancing performance requirements with the resources available. It can be an important factor to consider when implementing this approach in real-world settings. 

In the end, a proper and thorough evaluation of collaborative story generation remains an on-going research. While automatic evaluation metrics such as BERTScore has the best human correlations at story-level and system-level per ~\citet{chhun-etal-2022-human}, it may not be comprehensive enough in evaluating the highly creative output of collaborative story generation. There is a need for more nuanced and sophisticated metrics that can capture the complexity and diversity of collaborative stories. Therefore, the development and validation of appropriate evaluation methods is crucial for progress in this field.

\section{Ethical Considerations}
In Section~\ref{sec:data}, we have discussed our procedures to identify and remove potential harmful content and user privacy information. However, it is important to also consider the broader ethical implications of using AI in collaborative storytelling. These include issues such as ensuring fair and unbiased representation, protecting data privacy, and preventing the use of AI-generated content for harmful purposes. For example, AI-generated stories or characters may perpetuate stereotypes or reinforce societal biases if they are trained on biased data. Therefore, it is crucial to consider and address these ethical issues in order to create inclusive and responsible AI-generated stories that do not harm individuals or groups.

\bibliographystyle{acl_natbib}
\bibliography{anthology,custom}

\appendix

\newpage
\section{Appendix}

\subsection{Instruction Template examples}
\label{a1}

Please refer to Table~\ref{tab:instructions} for the instruction template examples.

\begin{table*}[]
    \centering
    \begin{tabular}{|p{4cm}|p{8cm}|p{3cm}|}
    \midrule
          task type & input format & output format  \\ 
          \midrule
          genre classification & \{story\} Is this a \{genre\} story? &  Yes or No \\
          \midrule
          authorship attribution & \{story\} Is this story written by \{author\}? &  Yes or No   \\
          \midrule
          authorship verification & Chapter A: \{chapter\textsubscript{a}\} Chapter B: \{chapter\textsubscript{b}\} Are the two story chapters above written by the same author? & Yes or No \\
          \midrule

          connectivity inference & Chapter A: \{chapter\textsubscript{a}\} Chapter B: \{chapter\textsubscript{b}\} Can Chapter B be the next chapter of Chapter A? & Yes or No \\
          \midrule
          temporal inference & Chapter A: \{chapter\textsubscript{a}\} Chapter B: \{chapter\textsubscript{b}\} Does Chapter A happen before Chapter B? & Yes or No \\
            \midrule
          story scoring & \{story\} How do you like the story above? Please rate the story from 0 to 10: & 0.0 - 10.0 \\
        \midrule
          story segmentation & \{story\} Please segment the story into chapters:  & \{c\textsubscript{1}\} ||| \{c\textsubscript{2}\} ||| \{c\textsubscript{3}\} ...   \\
         \midrule
        next chapter generation & \{story\textsubscript{0:i}\} Please write a next chapter for the above story:  &   \{chapter\textsubscript{i}\}  \\
         \midrule
         conditional story generation & \{story\textsubscript{0:i}\} Please finish the whole story:  & \{story\textsubscript{i:}\}   \\
         \midrule
         chapter infilling  & Chapter A: \{chapter\textsubscript{a}\} Chapter B: \{chapter\textsubscript{b}\}  Please write a chapter between Chapter A and Chapter B:   & \{chapter\textsubscript{i}\}   \\
         \midrule
         global infilling & Previous chapters: \{story\textsubscript{prev}\} Next chapters: \{story\textsubscript{next}\}  Based on the context of previous and next chapters, please fill in a chapter in between:  & \{chapter\textsubscript{i}\}   \\
         \midrule
         temporal ordering & \{story\textsubscript{permute}\} Please rewrite the story in correct temporal order:   &  \{story\textsubscript{correct}\}  \\
         \midrule

    \end{tabular}
    \caption{Instruction template examples.}
    \label{tab:instructions}
\end{table*}

\subsection{Hypterparameters}
\label{a2}

Please refer to Table~\ref{tab:params} for the hyperparameters.

\begin{table}[H]
	\small
    \centering
    \begin{tabular}{ccc}
    \midrule
          name & value  \\ 
          \midrule
          batch size & 64 \\
          learning rate & 5e-5 \\
          training steps & 50000 \\
          warmup steps & 2000 \\
         \midrule

    \end{tabular}
    \caption{Hypterparameters for \instruct}
    \label{tab:params}
\end{table}

\subsection{Full results tables}
\label{a3}

Please refer to Table~\ref{tab:fullmore}, Table~\ref{tab:fullmore2}, Table~\ref{tab:fewmore}, and Table~\ref{tab:zeromore} for all full results.

\begin{table*}[ht!]
	\small
    \centering
    \begin{tabular}{cccccc}
    \midrule
          task & \textbf{BERT} & \textbf{RoBERTa} & \textbf{DeBERTa} & \textbf{T5} & \textbf{InstructStory}  \\ 
         \midrule
\scriptsize  war  &  59.77  &  68.97  &  76.0  &  70.59  &  78.26 \\
\scriptsize  life  &  35.41  &  40.0  &  37.5  &  51.75  &  46.48 \\
\scriptsize  fanfiction  &  55.28  &  62.26  &  67.27  &  63.41  &  66.07 \\
\scriptsize  poetry  &  78.9  &  85.71  &  79.65  &  81.97  &  84.96 \\
\scriptsize  music  &  69.14  &  83.87  &  85.42  &  83.17  &  86.6 \\
\scriptsize  fantasy  &  43.7  &  47.37  &  48.75  &  47.95  &  50.98 \\
\scriptsize  humor  &  60.61  &  54.12  &  62.22  &  61.95  &  56.07 \\
\scriptsize  lgbt  &  48.08  &  60.24  &  63.83  &  59.81  &  55.77 \\
\scriptsize  school  &  36.14  &  63.24  &  65.22  &  51.22  &  51.76 \\
\scriptsize  game  &  58.62  &  77.55  &  77.42  &  68.24  &  69.57 \\
\scriptsize  sad  &  48.35  &  56.93  &  53.97  &  53.44  &  55.17 \\
\scriptsize  nature  &  39.51  &  51.43  &  48.08  &  51.85  &  47.17 \\
\scriptsize  magic  &  60.61  &  63.74  &  61.9  &  59.42  &  61.76 \\
\scriptsize  adventure  &  40.43  &  55.24  &  46.38  &  44.32  &  45.64 \\
\scriptsize  sci-fi  &  65.42  &  61.07  &  67.24  &  62.69  &  66.67 \\
\scriptsize  romance  &  54.84  &  59.68  &  60.29  &  56.52  &  62.12 \\
\scriptsize  hero  &  32.26  &  56.14  &  61.9  &  70.97  &  71.84 \\
\scriptsize  euphoric  &  28.26  &  40.35  &  44.83  &  44.59  &  43.1 \\
\scriptsize  space  &  72.73  &  74.23  &  78.72  &  80.0  &  78.9 \\
\scriptsize  survival  &  29.73  &  58.59  &  59.32  &  53.06  &  52.38 \\
\scriptsize  mystery  &  43.58  &  51.47  &  48.53  &  47.48  &  51.97 \\
\scriptsize  drama  &  42.67  &  45.3  &  46.43  &  44.21  &  47.4 \\
\scriptsize  royalty  &  72.73  &  74.0  &  68.18  &  74.75  &  75.47 \\
\scriptsize  dystopia  &  43.48  &  57.14  &  61.16  &  52.23  &  63.55 \\
\scriptsize  death  &  51.57  &  60.87  &  66.67  &  53.59  &  60.94 \\
\scriptsize  horror  &  45.67  &  55.64  &  60.15  &  52.05  &  53.33 \\
\scriptsize  animals  &  82.69  &  86.02  &  82.24  &  82.88  &  86.79 \\
\midrule
\scriptsize  intellikat  &  76.47  &  80.43  &  72.41  &  72.0  &  80.0 \\
\scriptsize  Hope!  &  61.82  &  81.13  &  62.3  &  56.21  &  68.22 \\
\scriptsize  ArtemisNine  &  46.58  &  68.42  &  58.14  &  65.98  &  69.09 \\
\scriptsize  Mockingjay  &  50.98  &  64.52  &  57.97  &  31.58  &  55.63 \\
\scriptsize  Rosetta  &  70.83  &  78.72  &  73.79  &  69.81  &  78.0 \\
\scriptsize  ember  &  46.6  &  68.09  &  59.26  &  55.71  &  55.12 \\
\scriptsize  CheshireinWonderland  &  47.31  &  55.42  &  63.04  &  40.7  &  58.41 \\
\scriptsize  Ellipsis  &  78.85  &  83.67  &  59.38  &  67.89  &  78.0 \\
\scriptsize  Scorpio :)  &  58.82  &  73.08  &  61.54  &  53.42  &  64.83 \\
\scriptsize  DANDAN THE DANDAN  &  63.27  &  70.73  &  76.6  &  65.22  &  71.11 \\
\scriptsize  Luke V.  &  72.09  &  69.77  &  69.23  &  63.24  &  73.79 \\
\scriptsize  Windlion  &  87.13  &  90.38  &  93.07  &  88.89  &  92.16 \\
\scriptsize  Kitin  &  86.87  &  83.72  &  78.18  &  80.0  &  74.42 \\
\scriptsize  Tricia L  &  43.84  &  70.09  &  61.29  &  45.59  &  64.71 \\
\scriptsize  Nathan.N  &  82.61  &  84.78  &  86.0  &  86.32  &  87.23 \\
\scriptsize  Zed  &  67.27  &  72.94  &  81.82  &  73.27  &  78.85 \\
\scriptsize  CAPSLOCK  &  77.59  &  74.38  &  80.81  &  67.96  &  80.37 \\
\scriptsize  R  &  65.26  &  88.89  &  85.71  &  78.26  &  88.89 \\
\scriptsize  go!den-in-the-mist  &  78.85  &  84.96  &  78.9  &  66.17  &  72.73 \\
\scriptsize  Libra ( inactive)  &  54.14  &  62.3  &  57.89  &  54.55  &  57.66 \\
\scriptsize  Silverfroststorm  &  75.79  &  67.83  &  55.7  &  51.5  &  63.16 \\
\scriptsize  Shasta  &  52.17  &  55.56  &  58.49  &  37.04  &  59.38 \\
\scriptsize  SaintSayaka  &  71.43  &  75.21  &  77.06  &  61.87  &  75.23 \\
\scriptsize  Amelia Rose  &  50.0  &  70.1  &  68.57  &  53.62  &  68.97 \\
\scriptsize  sagittarius  &  50.94  &  54.74  &  58.02  &  48.52  &  64.81 \\
\scriptsize  Phantim  &  66.67  &  81.55  &  78.1  &  70.59  &  76.79 \\
\scriptsize  Ara Argentum Aurum!  &  50.94  &  49.28  &  56.41  &  63.46  &  67.33 \\
\scriptsize  aspiringwriter  &  66.67  &  69.57  &  62.02  &  60.4  &  67.18 \\
\scriptsize  camel   &  71.15  &  73.12  &  77.06  &  64.41  &  66.67 \\
\scriptsize  darcy  &  62.65  &  65.98  &  63.64  &  66.67  &  64.86 \\
\midrule
\scriptsize author\_verification         & 23.19 & 23.41 & 23.17 & 22.94 & 23.57 \\ \midrule
\scriptsize temporal\_inference            & 72.90 & 77.74 & 80.18  & 78.51 & 79.04 \\ \midrule
\scriptsize connectivity\_inference           & 65.03 & 62.97 & 67.61 & 67.20  & 68.72 \\ \midrule
\scriptsize likes\_scoring                   & 53.54 & 75.74 & 60.81 & 67.35 &  68.82 \\
\scriptsize stars\_scoring                   & 55.34 & 66.60 & 56.02 & 63.15 &  63.26 \\ \midrule
 \scriptsize story\_segmentation           & 31.38 & 47.28 & 41.09 & 46.87 & 47.33 \\ \midrule

    \end{tabular}
    \caption{Fully-supervised understanding results of \instruct and other baselines.}
    \label{tab:fullmore}
\end{table*}

\begin{table*}[t]
	\small
    \centering
    \begin{tabular}{cccccc}
    \midrule
  Task & \textbf{GPT2-l} & \textbf{GPT2-m} & \textbf{OPT-350m} & \textbf{T5} & \textbf{InstructStory}  \\ 
\midrule

next\_chapter         & 81.35 & 80.90 & 83.25 & 82.17 & 82.43 \\ 
conditional      & 79.40 & 79.33 & 82.39 & 81.10 & 81.24 \\ 
chapter\_infilling                & 80.93 & 80.67 & 82.89 & 82.34 & 82.51 \\
 global\_infilling                   & 81.49 & 81.30 & 83.70 & 82.22 & 82.44 \\ 
 temporal\_ordering                 & 76.49 & 76.33 & 92.77 & 90.08  & 93.14 \\ 
\midrule

    \end{tabular}
    \caption{Fully-supervised generation results of \instruct and other baselines.}
    \label{tab:fullmore2}
\end{table*}

\begin{table*}[t]
	\small
    \centering
        \begin{tabular}{ccccccc}
    \midrule
         task & \textbf{BERT} & \textbf{RoBERTa} & \textbf{DeBERTa} & \textbf{T5} & \textbf{FLAN-T5} & \textbf{InstructStory}  \\ 
         \midrule
\scriptsize wordgames & 59.65 & 80.90 & 77.27 & 62.40 & 71.05 & 73.68 \\ 
                                \scriptsize rebellion         & 38.38 & 45.87 & 33.33 & 43.24 & 50.00 & 50.00 \\
                               \scriptsize mythology           & 47.27 & 59.79 & 61.54 & 62.07 & 66.67 & 67.33 \\ 
                               \scriptsize future         & 30.00 & 40.00 & 50.90 & 36.23 & 44.86 & 54.70 \\
                               \scriptsize friendship           & 38.82 & 46.96 & 44.62 & 49.23 & 53.33 & 55.36 \\
                               \scriptsize fairytale         & 45.93 & 60.32 & 65.52 & 74.07 & 72.09 & 79.59 \\
                               \scriptsize dreams           & 47.48 & 64.15 & 58.62 & 78.16 & 71.26 & 76.74 \\
                               \scriptsize crime         & 48.54 & 66.67 & 36.04 & 65.42 & 62.22 & 65.26 \\
                               \scriptsize change           & 44.00 & 50.36 & 32.91 & 33.90 & 47.89 & 39.19 \\
                               \scriptsize action         & 38.30 & 40.25 & 36.47 & 41.13 & 55.10 & 52.54 \\
\midrule
    \end{tabular}
    \caption{Few-shot results of \instruct and other baselines.}
    \label{tab:fewmore}
\end{table*}

\begin{table*}[t]
	\small
    \centering
    \begin{tabular}{cccc}
    \midrule
          task & \textbf{T5} & \textbf{FLAN-T5} & \textbf{InstructStory}  \\ 
         \midrule
\scriptsize disease  &  30.36  &  62.3  &  67.69 \\
\scriptsize harrypotter  &  29.63  &  84.21  &  85.71 \\
\scriptsize dragons  &  30.22  &  70.42  &  95.0 \\
\scriptsize art  &  34.53  &  54.84  &  87.36 \\
\scriptsize memories  &  32.65  &  40.0  &  70.18 \\
\scriptsize suspense  &  31.82  &  42.77  &  43.68 \\
\scriptsize supernatural  &  39.34  &  48.28  &  45.33 \\
\scriptsize angel  &  34.48  &  55.17  &  82.61 \\
\scriptsize revenge  &  35.0  &  58.06  &  57.14 \\
\scriptsize surreal  &  31.41  &  33.86  &  46.25 \\
\scriptsize history  &  38.6  &  54.12  &  60.34 \\
\scriptsize choices  &  40.51  &  28.7  &  50.0 \\
\scriptsize vampire  &  19.12  &  63.33  &  58.82 \\
\scriptsize lies  &  30.22  &  46.34  &  70.33 \\
\scriptsize crazy  &  30.0  &  42.31  &  43.08 \\
\scriptsize secret  &  36.19  &  39.49  &  44.59 \\
\scriptsize pirates  &  35.97  &  41.51  &  65.63 \\
\scriptsize world  &  30.63  &  34.92  &  50.75 \\
\scriptsize hope  &  36.99  &  38.6  &  57.14 \\
\scriptsize reality  &  32.56  &  39.56  &  39.47 \\
\scriptsize family  &  14.88  &  51.16  &  60.0 \\
\scriptsize emotions  &  34.67  &  34.67  &  60.18 \\
\scriptsize strange  &  28.19  &  34.55  &  38.64 \\
\midrule
    \end{tabular}
    \caption{Zero-shot results of \instruct and other baselines.}
    \label{tab:zeromore}
\end{table*}



\end{document}